\documentclass{article} 
\usepackage{iclr2023_conference,times}

\usepackage{amsmath,amsfonts,bm}

\def\eqref#1{equation~\ref{#1}}

\def\1{\bm{1}}

\def\eps{{\epsilon}}

\DeclareMathAlphabet{\mathsfit}{\encodingdefault}{\sfdefault}{m}{sl}
\SetMathAlphabet{\mathsfit}{bold}{\encodingdefault}{\sfdefault}{bx}{n}

\usepackage{hyperref}
\usepackage{url}
\usepackage{colortbl}
\usepackage{amsthm}

\usepackage[capitalize,noabbrev]{cleveref}

\theoremstyle{plain}
\newtheorem{theorem}{Theorem}[section]

\newtheorem{lemma}[theorem]{Lemma}

\theoremstyle{definition}
\newtheorem{definition}[theorem]{Definition}

\theoremstyle{remark}
\newtheorem{remark}[theorem]{Remark}

\usepackage{amssymb}
\usepackage{epsfig}
\usepackage{graphicx}
\usepackage{algorithm}
\usepackage{algpseudocode}
\usepackage{amsmath}
\usepackage{xcolor}
\usepackage{booktabs}
\usepackage{caption}
\usepackage{graphicx}
\usepackage{amsthm}
\usepackage{comment}
\usepackage{thmtools, thm-restate}
\usepackage{enumitem}

\numberwithin{theorem}{section}
\numberwithin{equation}{section}

\title{RACH-Space: Reconstructing Adaptive Convex Hull Space with Applications in Weak Supervision}

\author{%
  Woojoo Na \\
  \texttt{woojooya777@gmail.com} \\
  \AND
  Abiy Tasissa \\
  \texttt{abiy.tasissa@tufts.edu} \\
}

\usepackage{bm}
\def\A{\mathbf{A}}
\def\a{\mathbf{a}}

\def\b{\mathbf{b}}

\def\x{\mathbf{x}}
\def\y{\mathbf{y}}

\def\p{\mathbf{p}}

\def\X{\mathbf{X}}
\def\w{\mathbf{w}}
\def\W{\mathbf{W}}

\def\z{\mathbf{z}}
\def\eps{\bm{\epsilon}}

\newcommand{\conv}[1]{\textrm{Conv}(#1)}
\newcommand{\col}[1]{\textrm{Col}(#1)}
\DeclareMathOperator*{\minimize}{\mathrm{minimize}}

\iclrfinalcopy 
\begin{document}

\maketitle

\begin{abstract}
We introduce \emph{RACH-Space}, an algorithm for labelling unlabelled data in weakly supervised learning, given incomplete, noisy information about the labels. \emph{RACH-Space} offers simplicity in implementation without requiring hard assumptions on data or the sources of weak supervision, and is well suited for practical applications where fully labelled data is not available. Our method is built upon a geometrical interpretation of the space spanned by the set of weak signals. We also analyze the theoretical properties underlying the relationship between the convex hulls in this space and the accuracy of our output labels, bridging geometry with machine learning. Empirical results demonstrate that \emph{RACH-Space} works well in practice and compares favorably to the best existing label models for weakly supervised learning.

\end{abstract}

\section{Introduction}
Leveraging their effectiveness in handling extensive datasets, machine learning models are employed in various fields, such as image and video recognition \cite{dosovitskiy2020image,krizhevsky2012imagenet,haralick1973textural}, natural language processing \cite{wolf2020transformers}, and structure prediction \cite{varadi2022alphafold}. Despite these advancements, a major challenge remains: these models heavily depend on manually labelled training data \cite{bommasani2021opportunities}. The optimisation of deep learning models, often consisting of hundreds of millions of parameters, requires even larger amounts of labelled training data compared to traditional models \cite{bommasani2021opportunities}. Creating these datasets is both expensive and time-consuming. Additionally, these datasets are often tailored for specific purposes, making them less adaptable to new applications. As a result, the laborious and inflexible process of manually labelling datasets has emerged as a primary bottleneck in machine learning models \cite{nikolenko2021synthetic}.

To address the difficulty of obtaining labelled training data, numerous large-scale machine learning systems consider alternatives to supervised learning. These alternatives encompass methodologies like active learning \cite{settles1994active}, semi-supervised learning \cite{xiaojin2006semi}, transfer learning \cite{weiss2016survey}, and weakly supervised learning \cite{zhang2022survey}. Each of these approaches, motivated by the shared challenge of limited labelled training data, employs distinct strategies to overcome this obstacle. In this paper, our focus will be on weakly supervised learning.

The concept of weak supervision arises from scenarios where fully training data is too expensive or not available, but human supervision sources can be efficiently leveraged. This can be achieved by tapping into existing knowledge sources, such as a knowledge base \cite{mintz2009distant}, opting for more economical but lower-quality supervision, like crowd-sourcing \cite{dawid1979maximum}, or providing higher-level yet less precise supervision using heuristic rules \cite{shin2015incremental}. The origins of this method trace back to early crowd-sourcing research, where ground truth labels for unlabelled data are estimated using expectation maximization \cite{dawid1979maximum}.
Weakly supervised learning finds applications in various tasks, including computer vision \cite{chen2020weakly}, text classification \cite{chen2020weakly}, and sentiment analysis \cite{medlock2007weakly}. 

In weak supervision, complete labels for training data are unavailable. Instead, there is unlabelled data, along with access to one or more sources of weak supervision, represented by weak signals. Each weak signal has an accuracy, defined as the expected probability of the true label of the data, assumed to be in the range $[0,1]$. The primary technical challenge in weak supervision is consolidating and de-noising these diverse and often conflicting sources. Various approaches to weak supervision have been explored, including data programming \cite{ratner2016data}, which unifies weak signals into a single label to minimize expected loss, and learning with assumptions on data distribution \cite{kuang2022firebolt}. Other methods focus on constructing specific noise models or simultaneously learning these models during training \cite{ratner2018snorkel}.
\subsection{Overview of RACH-Space}
In our setup, we focus on the classification task for diverse forms of data, aiming to categorize $n$ data points represented in the data matrix $\X=[\x_1\, \x_2\,...\,\x_n]$, divided into into $k$ classes. Alongside the data points, we are given $m$ weak signals $\w_1,\w_2,...\w_m$ where $\w_i \in [0,1]^{nk}$. Notably, the $(k-1)n+i$-th entry of $\mathbf{\w}_i$ indicates the probability of the $i^{th}$ data point belonging to class $k$. Let $\W \in [0,1]^{m\times nk}$ denote the matrix with each rows corresponding to the $m$ weak signals. Our objective is to estimate the unknown ground-truth label, denoted by $\y \in \{0,1\}^{nk}$. In what follows, we provide a brief road-map that discusses how \emph{RACH-Space}, the proposed method in this paper, obtains this estimate.

\begin{figure}[h!]
    \centering
\includegraphics[scale=0.9]{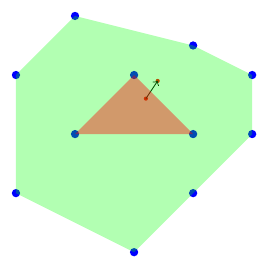}
    \caption{An illustration for the main idea of \emph{RACH-Space}. Using the upper bound $\frac{2}{k}-\frac{2}{k^2}$ for $\eps$, a right hand side vector $\widetilde{\b}$ is initialized. Our aim is to update the parameter $\eps$ so that $\frac{\widetilde{\b}}{n}$ lies in the \emph{safe region}, which is the region inside the convex hull but outside the second largest convex hull.}
    \label{fig:illustration}
\end{figure}

We begin by introducing the expected error rates of each weak signal $\w_i$, denoted as $\{\eps_i\}_{i=1}^{m}$. We can then formulate the label estimation problem to finding a solution to the linear system $\A\widetilde{\y} = \b$, where $\A = 2\W$, $\b \in \mathbb{R}^{m}$ with $b_i = -nk \cdot \eps_i +\mathbf{w}_i^{\top} \mathbf{1}+n$ and the constraint that $\widetilde{\y}\in [0,1]^{nk}$. Without making any prior assumptions about the weak signals or the unlabelled dataset, the individual expected error rates are unknown. Consequently, the entries of $\b$ are also unknown. Our algorithm, \emph{RACH-Space}, approaches this problem in two steps. First, we set up a framework to express the expected error rates $\eps_i$ using a single parameter $\eps$. Subsequently, we substitute each $\eps_i$ with this parameter to define $\widetilde{\b}$, an approximation for $\b$. Starting with this preliminary estimate, we iteratively refine it until $\frac{\widetilde{\b}}{n}$ falls within a region we refer to as the \emph{safe region}. Second, we estimate our ground truth label by solving a quadratic program. The \emph{safe region} is within the largest convex hull of the columns of $\A$ but outside the second largest convex hull of the columns of $\A$. The presence of $\frac{\widetilde{\b}}{n}$ in this region serves as implicit regularization that encourages the utilization of the parts of weak signals that indicates the most confidence in its corresponding data point's classification.
Figure \ref{fig:illustration} illustrates the main idea of \emph{RACH-Space}.
\paragraph{Contributions}
The contributions of this paper are:

1. We propose an algorithm, \emph{RACH-Space}, that consolidates multiple sources of partial, noisy information to effectively approximate the ground-truth label of unlabelled data.

2. We present a geometric framework for analyzing weak signals for weakly supervised learning. Based on the matrix of weak signals, we identify a set of nested convex hulls and  introduce the notion of a \emph{safe region}.

3. We present theoretical analysis of the utility of \emph{safe region} for representing the expected error rates of weak signals, which is unknown without prior knowledge of the data or the ground truth labels. We show empirical evidence of this utility on real world data.

4. We show that \emph{RACH-Space} performs at the state-of-the-art level performance compared to other weak supervision baselines over $14$ real-world datasets curated by the \emph{WRENCH} benchmark for weak supervision \cite{zhang2021wrench}.

\section{Related Work}
Most existing weak supervision models assume specific data or signal distributions to create synthetic labels \cite{zhang2022survey}. \cite{ratner2016data} uses a generative model, \cite{fu2020fast} assumes certain degree of class balance, and \cite{wu2023ground,yu2022learning,kuang2022firebolt} similarly utilize the assumptions to represent the distribution to formulate the synthetic label. Other methods like \cite{arachie2021constrained} and \cite{dawid1979maximum} assume a particular characteristic of the accuracy of the weak signals to produce the synthetic labels. Our method is philosophically similar to Constrained Label Learning \cite{arachie2021constrained} in the sense that we define a feasible space for $\widetilde{\y}$. However, our method is fundamentally different to \emph{Constrained Label Learning} \emph{(CLL)} in that \emph{RACH-Space} does not assume the prior knowledge of expected empirical rates of the weak signals. Whilst \emph{Hyper Label Model} \emph{(HLM)} \cite{wu2022learning} does not need an ad hoc dataset-specific parameter, it considers the setup where the entries of weak signals are one of $\{0, 1, -1\}$. 

\emph{RACH-Space} does not require the entries of weak signals to be an integer, and takes any input between $[0,1]$ (for label models in \emph{PWS}, it would be $[-1,1]$), thus allowing weak signals to express its confidence for each data points in terms of probability.

A pivotal aspect of \emph{RACH-Space} involves updating $\eps$ so that $\frac{\tilde{\b}}{n}$ lies within the \emph{safe region}, which is the area between the convex hull and the second largest convex hull. In our algorithm, this region assumes critical importance. We note that our approach is thematically related to the concept of convex layers or convex hull peeling, where a set of points are represented as a sequence of convex layers by progressively eliminating vertices from the convex hull \cite{calder2020limit, dalal2004counting, barnett1976ordering, chazelle1985convex}.

\section{RACH-Space}
\subsection{Problem Formulation}
Consider $n$ unlabelled training data points $\X=[\x_1, \x_2, ..., \x_n]$, with $m$ weak signals $\mathbf{w}_i$ constituting the set of weak signals $\W \in [0,1]^{m\times nk}$. These weak signals provide partial, often noisy information about classifying each data point into $k$ classes. The task is to find a label that provides the best approximation of the unknown ground truth label $\y \in \{0,1\}^{nk}$.

Table \ref{tab: weak_signals} is an illustration of weak signals along with the ground-truth label $\y \in \{0,1\}^{nk}$. This ground-truth label $\y$ is either unknown or too expensive to obtain in practice. In the typical setup, $m \ll nk$. Given weak signals $\W=[\mathbf{w}_1, . . . , \mathbf{w}_m]$, the objective of weakly supervised learning label models is to return the synthetic label $\widetilde{\y}$ which is an optimal estimate of $\y$. Thus, we consider minimizing the error between the estimate and true label for unlabelled dataset $\X=[\x_1, ... , \x_n]$.

\begin{table}[ht] 
\centering
\caption{Each weak signal of length $nk$ gives information about classifying $n$ data points into $k$ classes. The ground-truth label $\y^\top$ is unknown in practice. For data points where weak signals abstain ($\emptyset$) from making a prediction, we assign a probability of $\frac{1}{k}$.}\label{tab: weak_signals}
\resizebox{\linewidth}{!}{%
\begin{tabular}{p{\dimexpr\linewidth/10}|
                *{15}{p{\dimexpr(\linewidth-\linewidth/5)/15-2\tabcolsep-\arrayrulewidth}|}}
\cline{2-16}
                 & \multicolumn{5}{c|}{Class 1}    & \multicolumn{5}{c|}{Class2} & \multicolumn{5}{c|}{Class3} \\ 
\cline{2-16}
$\w_1$: & 0.8 & $\emptyset$ & 0.0 & 0.8 & 0.4 & $\emptyset$ & 0.7 & $\emptyset$ & 0.2 & $\emptyset$ & $\emptyset$ & $\emptyset$ & 0.6 & $\emptyset$ & $\emptyset$ \\
\cline{2-16}
$\w_2$: & 0.7 & 0.2 & $\emptyset$ & 0.6 & 0.3 & $\emptyset$ & $\emptyset$ & $\emptyset$ & $\emptyset$ & $\emptyset$ & $\emptyset$ & $\emptyset$ & $\emptyset$ & 0.3 & $\emptyset$ \\
\cline{2-16}
$\w_3$: & $\emptyset$ & $\emptyset$ & $\emptyset$ & $\emptyset$ & $\emptyset$ & 0.4 & $\emptyset$ & $\emptyset$ & 0.4 & 0.6 & $\emptyset$ & $\emptyset$ & $\emptyset$ & $\emptyset$ & 0.9 \\
\cline{2-16}
\end{tabular}
}
\noindent
\resizebox{\linewidth}{!}{%
\begin{tabular}{p{\dimexpr\linewidth/10}
                *{15}{p{\dimexpr(\linewidth-\linewidth/5)/15-2\tabcolsep-\arrayrulewidth}}}
Data: & $x_1$ & $x_2$ & $x_3$ & $x_4$ & $x_5$ & $x_1$ & $x_2$ & $x_3$ & $x_4$ & $x_5$ & $x_1$ & $x_2$ & $x_3$ & $x_4$ & $x_5$\\ 
\end{tabular}
}
\resizebox{\linewidth}{!}{%
\noindent
\begin{tabular}{p{\dimexpr\linewidth/10}|
                *{15}{p{\dimexpr(\linewidth-\linewidth/5)/15-2\tabcolsep-\arrayrulewidth}|}}
                \cline{2-16}
$\y^\top$ & 1 & 0 & 0 & 1 & 0 & 0 & 1 & 0 & 0 & 0 & 0 & 0 & 1 & 0 & 1   \\
\cline{2-16}
\end{tabular}
}
\end{table}

Now, we formally introduce the expected error rate $\eps_i$ for each weak signal. Here on, $\mathbf{1}$ indicates the $1$-vector. The expected error rate or weak signal $\w_i$, hereafter denoted as $\eps_i$ is given by:
\begin{equation}
\eps_i=\frac{1}{nk}( \mathbf{w_i}^\top (\mathbf{1}-\mathbf{y}) +  (\mathbf{1-w_i})^\top \mathbf{y}).
\end{equation}
As the ground-truth label $\y$ gives unique classification for each of the $n$ data points, i.e., $\mathbf{1}^\top \mathbf{y}=n$, the formula simplifies to:
\begin{equation} \label{eq:simple_expected_error}
\eps_i=\frac{1}{nk}( \mathbf{-2w_i}^\top \mathbf{y} + \mathbf{w_i}^\top \mathbf{1}+n).
\end{equation}

Since the ground-truth label $\y$ is unknown, we can not determine the individual expected error rates $\eps_i$ and, subsequently the vector $\b$. To address this challenge, we introduce a parameter $\eps$ as a ``reasonable" initial approximation for the individual expected error rates $\eps_i$. This parameter $\eps$ is then employed to define the variable $\widetilde{\b}$, which we will elaborate in the next section. Once $\widetilde{\b}$ is established, we delve into the detailed explanation of how we set the parameter $\eps$. A first natural choice
 of $\eps$ is the average expected error rate denoted by $\eps_{\text{avg}}$. However, akin to the individual expected error rates, direct access to $\eps_{\text{avg}}$ is not available without additional assumptions. Indeed, we argue that $\eps_{\text{avg}}$ may not even serve as a reliable representation of individual expected error rates $\eps_i$. To support this claim, we demonstrate that $\eps_{\text{avg}}$ can be artificially approximated to a specific asymptotic value by introducing particular types of weak signals.

With the definition of expected error rates $\eps_i$, we formally introduce the rest of the variables for \emph{RACH-Space}, which we use with \cref{eq:simple_expected_error} to form a quadratic objective. In addition to unlabelled training data $\X$, set of weak signals $\W$, and the expected error rate $\eps_i$ and parameter $\eps$ previously discussed, we use the following variables:

\begin{itemize}[leftmargin=*]

\item $\widetilde{\y} \in [0,1]^{nk}$, the output of \emph{RACH-Space}, our estimate for the ground-truth label $\y$.

\item $\A=2\W \in [0,2]^{m\times nk}$. This matrix follows from the simplified formulation in \cref{eq:simple_expected_error}.

\item $\b \in \mathbb{R}^{m}$ with 
$b_i = -nk \cdot \eps_i +\mathbf{w}_i^{\top} \mathbf{1}+n$. The $i^{th}$ entry is derived from individual error rates $\eps_i$ corresponding to the $i^{th}$ weak signal $\mathbf{w}_i$.

\item $\widetilde{\b} \in \mathbb{R}^m$ with $\widetilde{b}_i = -nk \cdot \eps + \mathbf{w_{i}}^\top \mathbf{1}+n$. Each row is defined in terms of the parameter $\eps$. 
\end{itemize}

Using \ref{eq:simple_expected_error}, we can design a least squares objective to obtain the synthetic label $\widetilde{\y}$. With that, the main quadratic objective of \emph{RACH-Space} can be now be formally stated as follows.
\begin{equation}
\label{eq:quadratic_objective}
    \minimize_{\widetilde{\y} \in [0,1]^{nk}} \quad||\A\widetilde{\y}-\widetilde{\b}||^2 \quad \text{subject to} \quad \mathbf{1}^ \top \widetilde{\y} = n.
\end{equation}

\subsection{Upper Bound on \texorpdfstring{$\eps$}{eps}}

\emph{RACH-Space} aims to capture the relation between individual expected error rates $\eps_i$, the weak signals and the unlabelled training data. We now describe how \emph{RACH-Space} chooses a ``reasonable'' representation $\eps$ of $\eps_i$.
First, a natural choice of such $\eps$ would be $\eps_{\text{avg}}$ which is unknown.
As discussed in the subsequent remark, the value of $\eps_{\text{avg}}$ can consistently be altered to an asymptotic value by introducing specific types of weak signals into the existing set of weak signals.\\

\begin{remark}
\label{rem: weak_signal_bound}
    By adding arbitrarily many $\w_{\text{random}}$ to the set of weak signals, we can always approximate the average expected error rates $\eps_{\text{avg}}$ of weak signals to be $\frac{2}{k}-\frac{2}{k^2}$. One can further decrease $\eps_{\text{avg}}$ of weak signals close to $\frac{1}{k}$ by adding $\w_{\text{null}}$ to the set of weak signals.
\end{remark}
In \cref{rem: weak_signal_bound}, $\w_{\text{random}} = \{1/k, ..., 1/k\}$ denotes a random classification of the data points, where $k$ is the number of classes. $\w_{\text{null}} = \{0, ..., 0\}$ indicates that each of the data points do not belong to any class. From our definition of expected error rates \ref{eq:simple_expected_error}, we have that the expected error rate $\eps_{random}$ of $\w_{\text{random}}$ is always $\frac{2}{k}-\frac{2}{k^2}$, and that the expected error rate $\eps_{null}$ of $\w_{\text{null}}$ is always $\frac{1}{k}$, regardless of the ground-truth label $\y$. Note, as $\frac{1}{k} \leq \frac{2}{k}-\frac{2}{k^2}$ for all class size $k$, the expected error rate of $\w_{\text{null}}$ is always better or equal to $\w_{\text{random}}$ and the strict equality holds for $k=2$, i.e. binary classification.

The procedure in \cref{rem: weak_signal_bound} of adding the two sets of weak signals $\w_{\text{random}}$ and $\w_{\text{null}}$ allows one to manipulate the set of weak signals to have a close approximation to the altered value of $\eps_{\text{avg}}$. In particular, we can always add $\w_{\text{random}}$ to stably approximate $\eps_{\text{avg}}$ to $\frac{2}{k}-\frac{2}{k^2}$. We can even approximate $\eps_{\text{avg}}$ to $\frac{1}{k}$ by adding $\w_{\text{null}}$ to the set of weak signals. Note, for multi-classification tasks where $k>2$, $\frac{2}{k}-\frac{2}{k^2}$ would thus be a strong upper bound that we can approximate $\eps_{\text{avg}}$ for as $\frac{1}{k} \leq \frac{2}{k}-\frac{2}{k^2}$. Also, all weak signals have a natural lower bound of $0$ corresponding to the case when the weak signal is actually the ground-truth label.

However, this procedure of adding $\w_{\text{random}}$ and $\w_{\text{null}}$ does not provide information that would positively impact the accuracy of the estimate $\widetilde{\y}$ of $\y$. Thus, the actual value of the initial $\eps_{\text{avg}}$ is not only unknown, but it is also not a reliable representation of $\eps_i$. Therefore, instead of using $\eps_{\text{avg}}$ as $\eps$, we only take this asymptotic value of $\frac{2}{k}-\frac{2}{k^2}$ as an upper bound for $\eps$. 

Now that we have established a ``reasonable'' upper bound of $\frac{2}{k}-\frac{2}{k^2}$ on $\eps$, we are going to describe another mechanism for choosing $\eps$, which updates its value starting from this upper bound. This following mechanism uses a geometric structure we identified, which we call \emph{safe region}.

\subsection{Using Convex Hull Structure to Update \texorpdfstring{$\eps$}{eps}}

Having established an upper bound of $\eps$, we establish a criterion for $\widetilde{\b}$, which is that $\frac{\widetilde{\b}}{n}$ needs to lie in the \emph{safe region}. As $\widetilde{\b}$ is defined in terms of $\eps$, this dictates how we update $\eps$. We begin with the following characteristic of our matrix $\A$.
\begin{remark}
\label{rm: strong_class_indication}
    Since $\A=2\W$, larger entries in $\A$ correspond to stronger class indication in $\W$. As all entries in $\A$ are in $[0,2]$, the extreme points of the set of columns in $\A$ correspond to parts of the weak signal $\W$ with the strongest class indication.
\end{remark}

Now, we formally introduce \emph{safe region}. As for the notations we use, given a set of vectors $\mathcal{S}=\{\mathbf{a}_1,\mathbf{a}_2,...,\mathbf{a}_r\}$ in $\mathcal{R}^{m}$, the convex hull of $\mathcal{S}$, denoted by $\conv{\mathcal{S}}$, is the set of all convex combinations of vectors in $\mathcal{S}$. $\col{\A}$ denotes the column space of $\A$.

\begin{definition}
    \textbf{Safe region} is the area in which $\frac{\widetilde{\b}}{n}$ needs to lie inside, so that \emph{RACH-Space} does not overlook the parts of weak signals in $\W$ with the strongest class indicators.
\end{definition}

In practice, the number of columns in matrix $\A$ is considerably greater than the column dimension i.e., $nk\gg m$. Consequently, there exists a suitable subset of these columns $\mathcal{H}_1$ that define the convex hull of $\col{\A}$. This gives rise to the characteristic of \emph{safe region} for $\A$ as the area inside $\conv{\mathcal{H}_1}$ but outside of $\conv{\mathcal{H}_2}$, where $\mathcal{H}_1$ is the set of columns of $\A$ that spans the convex hull of $\col{\A}$, and $\mathcal{H}_2$ is the set of remaining columns of $\A$.

We note that requiring $\frac{\widetilde{\b}}{n}$ to lie in the \emph{safe region} gives us another criterion for our selection of $\eps$. In particular, starting from the upper bound of $\frac{2}{k}-\frac{2}{k^2}$, our algorithm \emph{RACH-Space} chooses $\widetilde{\b}$ by updating $\eps$ so that $\frac{\widetilde{\b}}{n}$ lies in the \emph{safe region}. This entails decreasing $\eps$ by a fixed step size, stopping when $\frac{\widetilde{\b}}{n}$ lies in the \emph{safe region}.

\begin{remark}
\label{rm: ensure_out_of_H2}
    As all entries in $\A$ are in $[0,2]$, thus non-negative, decreasing $\eps$ by a fixed step size, which increases entries of $\frac{\widetilde{\b}}{n}$, will always ensure that $\frac{\widetilde{\b}}{n}$ is outside of $\conv{\mathcal{H}_2}$, if $\frac{\widetilde{\b}}{n}$ was initially inside $\conv{\mathcal{H}_2}$.
\end{remark}
\vspace{-0.5em}

We initialize $\eps$ as $\frac{2}{k}-\frac{2}{k^2}$ for two primary reasons. Firstly, it serves as an upper bound for $\eps$. Secondly, it is the value to which $\eps_{\text{avg}}$ can always be approximated to without the need for any prior knowledge or assumptions about the dataset. This value, derived from the expected error rate of a randomly generated classification signal $\w_{\text{random}}$, also acts as the default initial value for $\eps$, representing the environment where no prior knowledge of the dataset is available.

The goal now is to position $\frac{\widetilde{\b}}{n}$ within \emph{safe region}, defined within the convex hull that is formed by the weak signals. Note that, the convex hull structure spanned by the columns of $\A$, and thus the \emph{safe region} as well, is not significantly changed by adding weak signals $\w_{\text{random}}$ and $\w_{\text{null}}$. That is, choice of columns that make up $\mathcal{H}_1$ and $\mathcal{H}_2$ is not altered by adding weak signals $\w_{\text{random}}$ and $\w_{\text{null}}$. With the procedure in \cref{rem: weak_signal_bound}, $\eps_{\text{avg}}$ can be manipulated to have the same value $\frac{2}{k}-\frac{2}{k^2}$ for different sets of weak signals. However, the choice of $\eps$ updated with this criterion regarding \emph{safe region} is different, reflecting the unique information provided by each set of weak signals $\W$.

We note that once $\frac{\widetilde{\b}}{n}$ is within \emph{safe region}, we stop updating $\eps$. We then solve the quadratic program in \ref{eq:quadratic_objective} to obtain the synthetic label.

\section{Algorithm}

\begin{algorithm}[tb]
    \caption{RACH-Space}
    \label{alg:RACH-Space}
\begin{algorithmic}
  \State {\bfseries Input } Weak signals $[\w_1,...,\w_m]$
  \State $\A$ $\leftarrow$ $\A=2\W$
  \State $\widetilde{\b}$ $\leftarrow$ $\widetilde{b}_i = -nk \cdot \left(\frac{2}{k}-\frac{2}{k^2}\right) + \mathbf{w}_{i}^\top \mathbf{1}+n$ 
  \State $\widetilde{\y}$ $\leftarrow$ $\widetilde{\y} \sim U(0,1)^n$ 
  \State $\mathcal{H}_1$ is the set of columns that form the convex hull of $\A$
  \State $\mathcal{H}_2$ are the remaining columns, i.e., $\mathcal{H}_2 = \A \backslash \mathcal{H}_1$
  \While{$\frac{\widetilde{\b}}{n}$ $\in$ $\conv{\mathcal{H}_2}$}
  \State $\widetilde{\b}$ $\leftarrow$ $\widetilde{b}_i = -nk \cdot \left(\frac{2}{k}-\frac{2}{k^2}-\alpha \right) + \mathbf{w_{i}}^\top \mathbf{1}+n$
  \EndWhile
  \State Add a row of $1$'s to $\A$ and append $n$ at the bottom of $\widetilde{\b}$
  \While{$\widetilde{\y}$ not converged}
    \State Update $\widetilde{\y}$ with gradient descent by optimizing $f(\widetilde{\y})=||\A\widetilde{\y} - \widetilde{\b}||^2$
    \State Clip $\widetilde{\y}$ to $[0,1]^{nk}$
  \EndWhile
  \State {\bfseries Output } $\widetilde{\y}$
\end{algorithmic}
\end{algorithm}

See \cref{alg:RACH-Space} for the pseudo-code of our proposed algorithm \emph{RACH-Space}.  We start by initializing $\eps$ to $\frac{2}{k}-\frac{2}{k^2}$ and then update it by a small step size, $\alpha$. Since all the entries of $\A$ are in $[0,2]$, updating $\eps$ moves $\frac{\widetilde{\b}}{n}$ outside of $\conv{\mathcal{H}_2}$ by positively incrementing all entries of $\widetilde{\b}$. If $\widetilde{\b}$ is already outside of $\conv{\mathcal{H}_2}$, then the algorithm fixes $\eps$ as its initial value. This is where \emph{RACH-Space} updates $\widetilde{\b}$ into the \emph{safe region}. Finally, the algorithm uses gradient descent to optimize a modified least squares objective \ref{eq:quadratic_objective}, where the modification reflects the sum to one constraint to the least squares objective.

\subsection{Time Complexity \& Adaptations}
A central part of \emph{RACH-Space} is its decision of $\eps$ through the convex hull structure of $\A$. This first requires the identification of the set of column vectors $\mathcal{H}_1$, from which the remaining columns $\mathcal{H}_2$ are naturally derived. Then, \emph{RACH-Space} checks for each update of $\eps$ whether $\frac{\widetilde{\b}}{n}$ is in $\conv{\mathcal{H}_2}$ or not. We use Qhull \cite{barber1996quickhull} to identify $\mathcal{H}_1$. This process becomes resource-intensive for large $m$, meaning when $\col{\A}$'s dimension is big. Specifically, the computational cost is $O((nk)^{\left \lfloor{\frac{m}{2}}\right \rfloor})$ \cite{barber1996quickhull}. This is $O(n)$ for $m=2,3$ and $O(n^2)$ for $m=4,5$. For experiments on real-world datasets, make this runtime reasonable by reducing the number of weak signals. We do this by aggregating the weak signals by dividing them into groups and deriving a new weak signal by average each entry of the group. Aggregating weak signals this way did not have a negative impact on the performance of \emph{RACH-Space}, as it did not significantly impact the convergence of $\widetilde{\y}$ computed from \emph{RACH-Space} when different group sizes were chosen. Checking whether $\frac{\widetilde{\b}}{n}$ is in $\conv{\mathcal{H}_2}$ can be readily completed by certifying whether the associated linear program is feasible. After $\eps$ is decided through this process, the quadratic optimisation problem \ref{eq:quadratic_objective} quickly converges to a unique solution $\widetilde{\y}$ after few iterations of gradient descent.

\section{Safe Region}

The main idea of the \emph{safe region} is to ensure that $\widetilde{\y}$, the output label of \emph{RACH-Space}, does not overlook the parts of weak signals $\W$ with the strongest class indication. In this section, we show that this corresponds to the area between $\conv{\mathcal{H}_1}$ and $\conv{\mathcal{H}_2}$.

\subsection{Solution Space inside \texorpdfstring{$\conv{\mathcal{H}_1}$}{conv(H1)}}

By construction, $\b$ meets the condition $\A\y=\b$, where $\mathbf{1}^\top \y = n$ and $\y \in {0,1}^{nk}$, representing an unknown ground-truth label $\y$. This implies that $\b$ is a convex combination of the columns $\col{\A}$ of matrix $\A$, specifically $\frac{\b}{n} \in \conv{\col{\A}}$. To find a suitable representation $\eps$ for $\eps_i$'s, we incorporate this as an additional criterion for $\eps$, requiring that $\widetilde{\b}$ defined by $\eps$ adheres to the condition $\frac{\widetilde{\b}}{n} \in \conv{\mathcal{H}_1}$, where we substitute $\col{\A}$ with its equivalent $\conv{\mathcal{H}_1}$. It is important to note that we employ a flexible entry for synthetic labels $\widetilde{\y}$, allowing $\widetilde{\y} \in [0,1]^{nk}$. Now, we show that relaxing this entry constraint for $\widetilde{\y}$ in \ref{eq:quadratic_objective} with a non-negative constraint admits infinitely many solutions for $\widetilde{\y}$ if $\frac{\widetilde{\b}}{n} \in \conv{\mathcal{H}_1}$. We begin by writing this relaxation as the following quadratic program:
\begin{equation}
\label{eq:least_squares_b}
    \min_{\widetilde{\y} \ge \mathbf{0}} \quad\bigg|\bigg|\A\widetilde{\y}-\widetilde{\b}\bigg|\bigg|^2 \quad \text{subject to} \quad \mathbf{1}^{\top}\widetilde{\y} = n.
\end{equation} 
In the following theorem, we show that if $\frac{\widetilde{\b}}{n} \in \conv{\col{\A}}$, this problem \ref{eq:least_squares_b} admits infinitely many solutions $\widetilde{\y}$.
\begin{theorem} \label{thm:infinite_solution_thm}
For $\frac{\widetilde{\b}}{n} \in \conv{\col{\A}}$ and $nk>m+1$, the optimisation program in \ref{eq:least_squares_b} has infinitely many solutions.
\end{theorem}

The proof of this theorem relies on the notion of affine independence defined below. 
\begin{definition}
    The vectors $\p_1,\p_2,...,\p_k$ are affinely independent if the vectors $\p_2-\p_1,\p_3-\p_1,...,\p_k-\p_1$ are linearly independent. Otherwise, the vectors are affine dependent.
\end{definition}
\begin{proof}
This proof is similar to the standard proof used to establish Carathéodory's theorem in convex analysis.  
Consider the vectors $\a_2-\a_1,\a_3-\a_1,...,\a_{nk}-\a_1$ where $\a_i$ denotes the $i^{th}$ column of $\A$. Since $nk>m+1$, these set of vectors are affinely dependent. This implies that there exists constants $d_2,d_3,..,d_{nk}$, which are all not zero, such that $\sum_{i=2}^{nk} d_i(\a_i-\a_1)=\mathbf{0}$. Equivalently, this can be written as $\sum_{i=1}^{nk}d_i\a_i=\mathbf{0}$ where $d_1=-\sum_{i=2}^{nk}d_i$. For latter use, note that $\sum_{i=1}^{nk}d_i=0$. Using the fact that $\frac{\widetilde{\b}}{n} \in \conv{\col{\A}}$, there exists a $\widetilde{\mathbf{y}}\ge \mathbf{0}$ with $\mathbf{1}^{\top}\widetilde{\mathbf{y}}=n$ such that $\A\mathbf{\widetilde{y}}=\widetilde{\b}$. We now propose a new representation of $\widetilde{\b}$ as follows: $\widetilde{\b} = 
\sum_{i=1}^{nk} (\widetilde{y}_i+\alpha d_i)\a_i$ where $\alpha>0$. For this to lead to another optimal solution, it suffices to check that $\widetilde{y}_i+\alpha d_i\ge 0$ for all values of $i$. For $d_i\ge 0$, $\widetilde{y}_i+\alpha d_i\ge 0$ holds automatically.  The relevant cases are the values of $i$ for which $d_i<0$. Setting $\alpha^* = \underset{i:d_i<0}{\min}\,-\frac{\widetilde{y}_i}{d_i}$, we see that $\widetilde{y}_i+\alpha^* d_i\ge 0$ and $\sum_{i=1}^{nk} (\widetilde{y}_i +\alpha^*d_i) = n$. Therefore, we have a new solution $\widetilde{\mathbf{z}}$ which differs from $\widetilde{\mathbf{y}}$. Once we have this pair of solutions, we can generate infinitely many solutions using $\widetilde{\mathbf{w}} = \lambda \widetilde{\mathbf{y}} + (1-\lambda)\widetilde{\mathbf{z}}$ for any $\lambda \in (0,1)$.   
\end{proof}

\cref{thm:infinite_solution_thm} illustrates that when $\frac{\widetilde{\b}}{n} \in \conv{\mathcal{H}_1}$, there are infinitely many convex combinations that reconstruct it. In what follows, we discuss how to further restrict the solution space inside $\conv{\mathcal{H}_1}$,

\subsection{Restricting the Solution Space inside \texorpdfstring{$\conv{\mathcal{H}_1}$}{conv(H1)}}

Now, we illustrate how we further restrict the space inside $\conv{\mathcal{H}_1}$ that we want $\frac{\widetilde{\b}}{n}$ to lie in. In particular, we describe why we want $\frac{\widetilde{\b}}{n} \notin$ $\conv{\mathcal{H}_2}$. This completes our description for \emph{safe region}, and why we update $\eps$ so that $\frac{\widetilde{\b}}{n} \notin$ $\conv{\mathcal{H}_2}$. 

\begin{lemma} \label{lem: cannot_converge_to_complete_wrong_label}
   If $\frac{\widetilde{\b}}{n} \notin$ $\conv{\mathcal{H}_2}$, the computed synthetic label $\widetilde{\y}$ cannot converge to a label where all the entries corresponding to the extreme points of $\conv{\col{\A}}$ are labelled $0$.
\end{lemma}

\begin{proof}    
Suppose the computed synthetic label converged to a label where all the entries corresponding to the extreme points of $\conv{\mathcal{H}_1}$ are labelled $0$. Let's call this label $\z$. Then by construction, from this label $\z$ we can derive a convex combination of the columns of $\mathcal{H}_2$ that is equivalent to $\frac{\widetilde{\b}}{n}$. Therefore we have that $\frac{\widetilde{\b}}{n} \in$ $\conv{\mathcal{H}_2}$.
\end{proof}

\cref{lem: cannot_converge_to_complete_wrong_label} shows that placing $\frac{\widetilde{\b}}{n}$ out of $\conv{\mathcal{H}_2}$ effectively prevents the quadratic program \ref{eq:quadratic_objective} from converging to $0$ at the extreme points of \conv{\col{$\A$}}. By \cref{rm: strong_class_indication}, this prevents \emph{RACH-Space} from contradicting the parts of weak signal $\W$ with the strongest class indication, whether that indicates inclusion or exclusion of a data point in a class. The significance of this procedure depends on how correct the strongest indicators are, and how much portion of the columns of $\A$ constitute $\mathcal{H}_1$. This is illustrated on real-world data in Section 6.2, which compares the case $\frac{\widetilde{\b}}{n} \notin$ $\conv{\mathcal{H}_2}$ and $\frac{\widetilde{\b}}{n} \in$ $\conv{\mathcal{H}_2}$ through the accuracy of the output label $\widetilde{\y}$.

\section{Experiments}
We evaluate our proposed method on $14$ real-world datasets curated by the \emph{WRENCH} weak supervision benchmark \cite{zhang2021wrench}. These datasets are in accordance with the \emph{Programmatic weak supervision} \emph{(PWS)} \cite{ratner2016data}. In \emph{PWS}, labelling functions (LFs) process data points and output a noisy label. Thus, LFs act as a form of weak supervision and are considered weak signals. All LFs in \emph{WRENCH} are from the original authors of each dataset \cite{zhang2021wrench}. The LFs yield weak signals with entries in $\{-1, 0, +1\}$, where $+1$ and $0$ signify class membership, and $-1$ indicates abstention. \emph{RACH-Space} converts format to have entries in $\{ \emptyset, 0, 1\}$ where $\emptyset$ indicates abstention. We highlight that this is a slightly modified setup compared to the setup for \emph{RACH-Space} which takes broader inputs as weak signals where each entry can take any values in $\{ \emptyset, [0, 1]\}$, allowing room for weak signals to indicate class membership in terms of probability. We only use this slightly limited expression for the sake of comparison to other baselines with the datasets in \emph{WRENCH} benchmark. 

\begin{table}[t]
\centering
\caption{$14$ real-world datasets curated by \emph{WRENCH} benchmark for weakly supervised learning \cite{zhang2021wrench}}  \label{tab: data}
\resizebox{\linewidth}{!}{%
\begin{tabular}{lllllllllllllll}
    \toprule
    Dataset & Census & Yelp & Youtube & CDR & Basketball & AGNews & TREC & SemEval & ChemProt & Spouse & IMDb & Commercial & Tennis & SMS\\ \midrule
    Task & income & sentiment & spam & bio relation & video frame & topic & question & relation & chemical relation & relation & sentiment & video frame & video frame & spam\\ \midrule
    \#class & $2$ & $2$ & $2$ & $2$ & $2$ & $4$ & $6$ & $9$ & $10$ & $2$ & $2$ & $2$ & $2$ & $2$ \\ \midrule
    metric & F1 & acc & acc & F1 & F1 & acc & acc & acc & acc & F1 & acc & F1 & F1 & F1\\ \midrule
    \#LF & $83$ & $8$ & $10$ & $33$ & $4$ & $9$ & $68$ & $164$ & $26$ & $9$ & $8$ & $4$ & $6$ & $73$\\ \midrule
    \#train & $10083$ & $30400$ & $1586$ & $8430$ & $17970$ & $96000$ & $4965$ & $1749$ & $12861$ & $22254$ & $20000$ & $64130$ & $6959$ & $4571$ \\ \midrule
    \#validation & $5561$ & $3800$ & $120$ & $920$ & $1064$ & $12000$ & $500$ & $178$ & $1607$ & $2801$ & $2500$ & $9479$ & $746$ & $500$\\ 
    \midrule
    \#test & $16281$ & $3800$ & $250$ & $4673$ & $1222$ & $12000$ & $500$ & $600$ & $1607$ & $2701$ & $2500$ & $7496$ & $1098$ & $500$\\ \bottomrule
\end{tabular}%
}
\end{table}

\subsection{Experiment 1: Comparison to Other Baselines}
We tested our algorithm on $14$ real-world datasets curated by the benchmark \cite{zhang2021wrench}, which represents diverse classification tasks weakly supervised learning can be applied to in the real world. We conducted our experiments using benchmark metrics \cite{zhang2021wrench}, averaging each metric over five runs for consistency. \cref{tab: data} shows the statistics of each dataset.

\subsection{Experiment: Performance of label models}

Our empirical experiments were conducted using the metrics provided by the benchmark ~\citep{zhang2021wrench} for each dataset, where each metric value is averaged over $5$ runs. Our experiments are conducted on $14$ datasets on \emph{WRENCH} benchmark, which covers various classification tasks and includes multi-class classification. Table 2 shows the statistics of each dataset.

\textbf{Label models:} $(1)$ \emph{Majority voting} (\emph{MV}). For each point, the predicted label is determined by choosing the most common label provided by the $m$ weak signals. Formally, it would simply count the number of $1$'s and $0$'s in the $m$ weak signals for the corresponding data point and choose whichever is more common as the data point's predicted label. $(2)$ \emph{Dawid-Skene} (\emph{DS}) \cite{dawid1979maximum} assumes a naive Bayes distribution over the weak signals and the ground-truth label to estimate the accuracy of each weak signal. $(3)$ \emph{Data Programming} (\emph{DP}) \cite{ratner2016data} describes the distribution of $p(L,Y)$ as a factor graph, where $L$ is the LF and $Y$ is the ground-truth label. $(4)$ \emph{MeTaL} \cite{ratner2019training} models $p(L,Y)$ via Markov Network, and \cite{ratner2018snorkel} uses it for weak supervision. $(5)$ \emph{FlyingSquid} (\emph{FS}) \cite{fu2020fast} models $p(L,Y)$ as a binary Ising model and requires label prior. It is designed for binary classification but one-versus-all reduction method was applied for multi-class classification tasks. $(6)$ \emph{Constrained Label Learning} (\emph{CLL}) \cite{arachie2021constrained} requires prior knowledge of the expected empirical rates for each weak signal to compute a constrained space from which they randomly select the synthetic labels from. For our experiments, we ran \emph{CLL} with the assumption that all weak labels are better than random. $(7)$ \emph{Hyper Label Model} (\emph{HLM}) \cite{wu2022learning} trains the model on synthetically generated data instead of actual datasets. Note that the difference in our experiment results from \cite{wu2022learning} is because their experiments were conducted in transductive setting \cite{mazzetto2021adversarial}, where data points used in learning is also used to evaluate the learned model. Hence their experiments are done where the train, validation and test datasets are merged for the label models to learn and to be evaluated. Our experiments adhere to the \emph{WRENCH} benchmark's original setup \cite{zhang2021wrench}, training label models on training data and evaluating them on test data.

\textbf{Results:} In our comparison in \cref{tab: experiment1}, we include all baselines that showed best performance in at least one dataset in the \emph{WRENCH} benchmark. \emph{RACH-Space} outperforms all other existing models, including the previous best-performing baseline by $0.47$ points. We used reduced the number of weak signals by dividing them into $5$ groups in given order and averaging them, and used a step size of $\alpha = 0.003$. For the \emph{IMDb} dataset, we further reduced the number of groups to four. During this process, when there are more than $5$ weak signals for the dataset, the entries are no longer integers. Since \emph{RACH-Space} does not assume an integer entry for weak signals, this is not a problem for \emph{RACH-Space}. This is also why we did not conduct additional experiments outside of the \emph{WRENCH} framework where weak signals can have fractional inputs. Our results are consistent with the benchmark \cite{zhang2021wrench}, with accuracy differences attributed to the inherent randomness in non-deterministic models.

\begin{table}[t]
\centering
\caption{Performance on all $14$ real-world datasets curated by \emph{WRENCH} benchmark\cite{zhang2021wrench}}  \label{tab: experiment1}
\resizebox{\linewidth}{!}{%
\begin{tabular}{llllllllllllllllllll}
    \toprule
    Dataset & Census & Yelp & Youtube & CDR & Basketball & AGNews & TREC & SemEval & ChemProt & Spouse & IMDb & Commercial & Tennis & SMS & AVG. \\ \midrule
    
    MV & 32.80 & 70.21 & 84.00 & 60.31 & 16.33 & 63.84 & 60.80 & 77.33 & \textcolor{red}{\textbf{49.04}} & 20.80 & \textcolor{red}{\textbf{71.04}} & \textcolor{red}{\textbf{85.28}} & 81.00 & \textcolor{blue}{\textbf{23.97}} & 56.91\\ \midrule
    
    DS & 47.16 & \textcolor{red}{\textbf{71.45}} & 83.20 & 50.43 & 13.79 & 62.76 & 50.00 & 71.00 & 37.59 & 15.53 & 70.60 & \textcolor{blue}{\textbf{88.24}} & 80.65 & 4.94 & 53.38\\ \midrule
    
    DP & 12.60 & 69.38 & 82.24 & 55.12 & 17.39 & \textcolor{red}{\textbf{63.95}} & 63.28 & 71.00 & 46.86 & 21.21 & 70.96 & 77.28 & \textcolor{blue}{\textbf{83.14}} & \textcolor{red}{\textbf{23.96}} & 54.16\\ \midrule
    
    MeTaL & 38.12 & 55.29 & 60.40 & 30.80 & 0 & 64.18 & 50.48 & 54.17 & \textcolor{blue}{\textbf{49.84}} & 20.13 & 69.96 & 77.95 & 80.54 & 23.83 & 48.26\\ \midrule
    
    FS & 17.77 & \textcolor{blue}{\textbf{71.68}} & 78.40 & \textcolor{red}{\textbf{63.22}} & 17.39 & 63.55 & 57.80 & 12.50 & 46.55 & 21.14 & 70.36 & 81.84 & 77.79 & 23.86 & 50.28\\ \midrule
    
    CLL & 26.99 & 51.89 & 57.60 & 24.97 & 16.12 & \textcolor{blue}{\textbf{64.83}} & 61.24 & 78.83 & 46.79 & \textcolor{red}{\textbf{22.56}} & 49.52 & 38.32 & 25.47 & 15.34 & 41.46\\ \midrule
    
    HLM & \textcolor{blue}{\textbf{56.31}} & 69.42 & \textcolor{red}{\textbf{85.60}} & 60.63 & \textcolor{red}{\textbf{17.60}} & 63.75 & \textcolor{blue}{\textbf{66.20}} & \textcolor{blue}{\textbf{82.50}} & 46.73 & 20.80 & \textcolor{blue}{\textbf{71.84}} & 82.76 & \textcolor{red}{\textbf{82.44}} & 23.19 & \textcolor{red}{\textbf{59.27}}\\ \midrule
    
    RACH-Space & \textcolor{red}{\textbf{52.82}} & 67.32 & \textcolor{blue}{\textbf{87.60}} & \textcolor{blue}{\textbf{69.98}} & \textcolor{blue}{\textbf{17.66}} & 62.63 & \textcolor{red}{\textbf{63.80}} & \textcolor{red}{\textbf{80.17}} & 45.12 & \textcolor{blue}{\textbf{44.51}} & 69.60 & 78.28 & 73.28 & 23.63 & \textcolor{blue}{\textbf{59.74}}\\
    \bottomrule
\end{tabular}%
}
\end{table}

\subsection{Experiment 2: Utility of Safe Region}

On the same real-world datasets curated by the \emph{WRENCH} benchmark, we compare the case $\frac{\widetilde{\b}}{n} \notin$ $\conv{\mathcal{H}_2}$ and $\frac{\widetilde{\b}}{n} \in$ $\conv{\mathcal{H}_2}$ through the accuracy of the output label $\widetilde{\y}$. We use the same setup for Experiment $1$, but instead reverse the updating rule for $\eps$ to place $\frac{\widetilde{\b}}{n}$ inside $\conv{\mathcal{H}_2}$. Results are summarized in \cref{tab: experiment2}, illustrating different empirical significance of \emph{safe region} depending on the dataset.

\begin{table}[t]
\centering
\caption{Performance comparison when $\frac{\widetilde{\b}}{n}$ outside vs inside of \emph{safe region}} \label{tab: experiment2}
\resizebox{\linewidth}{!}{%
\begin{tabular}{lllllllllllllllllll}
    \toprule
    Dataset & Census & Yelp & Youtube & CDR & Basketball & AGNews & TREC & SemEval & ChemProt & Spouse & Imdb & Commercial & Tennis & SMS \\ \midrule
    $\frac{\widetilde{\b}}{n} \in$ $\conv{\mathcal{H}_2}$ & 38.22 & 64.21 & 82.80 & 69.51 & 17.53 & 10.39 & 12.60 & 2.78 & 5.54 & 44.51 & 57.03 & 66.11 & 36.24 & 23.48 \\ \midrule
    
    $\frac{\widetilde{\b}}{n} \notin$ $\conv{\mathcal{H}_2}$ & 52.82 & 67.32 & 87.60 & 69.98 & 17.66 & 62.63 & 63.80 & 80.17 & 45.12 & 44.51 & 69.60 & 78.28 & 73.28 & 23.63 \\ \bottomrule
\end{tabular}%
}
\end{table}

\section{Conclusion}
We present \emph{RACH-Space}, a novel algorithm for labelling data under weakly supervised learning with only incomplete, noisy label information. By describing each source of information (weak signals) as a matrix, we identify a particular geometric structure inherent in the column space of this matrix, which are the nested set of convex hulls. After identifying this convex hull structure, we connect the geometric structure to a reliable representation of how bad each weak signal can be. We do this by deriving a singular value representation for the set of expected error rates of the weak signals. Thus, our geometric analysis aids in making a reasonable yet effective selection for this representation. Our approach demonstrates competitive performance compared to existing baselines on real-world datasets curated by weak supervision benchmark framework. Overall, we found \emph{RACH-Space} simple yet robust, well suited to a wide range of classification tasks where fully labelled data is unavailable.


\section*{Impact Statements}
This paper presents work whose goal is to advance the field of Machine Learning. There are many potential societal consequences of our work, none which we feel must be specifically highlighted here.

\bibliography{iclr2023_conference}

\begin{thebibliography}{32}
\providecommand{\natexlab}[1]{#1}
\providecommand{\url}[1]{\texttt{#1}}
\expandafter\ifx\csname urlstyle\endcsname\relax
  \providecommand{\doi}[1]{doi: #1}\else
  \providecommand{\doi}{doi: \begingroup \urlstyle{rm}\Url}\fi

\bibitem[Arachie \& Huang(2021)Arachie and Huang]{arachie2021constrained}
Chidubem Arachie and Bert Huang.
\newblock Constrained labeling for weakly supervised learning.
\newblock In \emph{Uncertainty in Artificial Intelligence}, pp.\  236--246. PMLR, 2021.

\bibitem[Barber et~al.(1996)Barber, Dobkin, and Huhdanpaa]{barber1996quickhull}
C~Bradford Barber, David~P Dobkin, and Hannu Huhdanpaa.
\newblock The quickhull algorithm for convex hulls.
\newblock \emph{ACM Transactions on Mathematical Software (TOMS)}, 22\penalty0 (4):\penalty0 469--483, 1996.

\bibitem[Barnett(1976)]{barnett1976ordering}
Vic Barnett.
\newblock The ordering of multivariate data.
\newblock \emph{Journal of the Royal Statistical Society: Series A (General)}, 139\penalty0 (3):\penalty0 318--344, 1976.

\bibitem[Bommasani et~al.(2021)Bommasani, Hudson, Adeli, Altman, Arora, von Arx, Bernstein, Bohg, Bosselut, Brunskill, et~al.]{bommasani2021opportunities}
Rishi Bommasani, Drew~A Hudson, Ehsan Adeli, Russ Altman, Simran Arora, Sydney von Arx, Michael~S Bernstein, Jeannette Bohg, Antoine Bosselut, Emma Brunskill, et~al.
\newblock On the opportunities and risks of foundation models.
\newblock \emph{arXiv preprint arXiv:2108.07258}, 2021.

\bibitem[Calder \& Smart(2020)Calder and Smart]{calder2020limit}
Jeff Calder and Charles~K Smart.
\newblock The limit shape of convex hull peeling.
\newblock \emph{Duke Mathematical Journal}, 169\penalty0 (11):\penalty0 2079--2124, 2020.

\bibitem[Chazelle(1985)]{chazelle1985convex}
Bernard Chazelle.
\newblock On the convex layers of a planar set.
\newblock \emph{IEEE Transactions on Information Theory}, 31\penalty0 (4):\penalty0 509--517, 1985.

\bibitem[Chen \& Batmanghelich(2020)Chen and Batmanghelich]{chen2020weakly}
Junxiang Chen and Kayhan Batmanghelich.
\newblock Weakly supervised disentanglement by pairwise similarities.
\newblock In \emph{Proceedings of the AAAI Conference on Artificial Intelligence}, volume~34, pp.\  3495--3502, 2020.

\bibitem[Dalal(2004)]{dalal2004counting}
Ketan Dalal.
\newblock Counting the onion.
\newblock \emph{Random Structures \& Algorithms}, 24\penalty0 (2):\penalty0 155--165, 2004.

\bibitem[Dawid \& Skene(1979)Dawid and Skene]{dawid1979maximum}
Alexander~Philip Dawid and Allan~M Skene.
\newblock Maximum likelihood estimation of observer error-rates using the em algorithm.
\newblock \emph{Journal of the Royal Statistical Society: Series C (Applied Statistics)}, 28\penalty0 (1):\penalty0 20--28, 1979.

\bibitem[Dosovitskiy et~al.(2020)Dosovitskiy, Beyer, Kolesnikov, Weissenborn, Zhai, Unterthiner, Dehghani, Minderer, Heigold, Gelly, et~al.]{dosovitskiy2020image}
Alexey Dosovitskiy, Lucas Beyer, Alexander Kolesnikov, Dirk Weissenborn, Xiaohua Zhai, Thomas Unterthiner, Mostafa Dehghani, Matthias Minderer, Georg Heigold, Sylvain Gelly, et~al.
\newblock An image is worth 16x16 words: Transformers for image recognition at scale.
\newblock \emph{arXiv preprint arXiv:2010.11929}, 2020.

\bibitem[Fu et~al.(2020)Fu, Chen, Sala, Hooper, Fatahalian, and R{\'e}]{fu2020fast}
Daniel Fu, Mayee Chen, Frederic Sala, Sarah Hooper, Kayvon Fatahalian, and Christopher R{\'e}.
\newblock Fast and three-rious: Speeding up weak supervision with triplet methods.
\newblock In \emph{International Conference on Machine Learning}, pp.\  3280--3291. PMLR, 2020.

\bibitem[Haralick et~al.(1973)Haralick, Shanmugam, and Dinstein]{haralick1973textural}
Robert~M Haralick, Karthikeyan Shanmugam, and Its'~Hak Dinstein.
\newblock Textural features for image classification.
\newblock \emph{IEEE Transactions on systems, man, and cybernetics}, pp.\  610--621, 1973.

\bibitem[Krizhevsky et~al.(2012)Krizhevsky, Sutskever, and Hinton]{krizhevsky2012imagenet}
Alex Krizhevsky, Ilya Sutskever, and Geoffrey~E Hinton.
\newblock Imagenet classification with deep convolutional neural networks.
\newblock \emph{Advances in neural information processing systems}, 25, 2012.

\bibitem[Kuang et~al.(2022)Kuang, Arachie, Liang, Narayana, DeSalvo, Quinn, Huang, Downs, and Yang]{kuang2022firebolt}
Zhaobin Kuang, Chidubem~G Arachie, Bangyong Liang, Pradyumna Narayana, Giulia DeSalvo, Michael~S Quinn, Bert Huang, Geoffrey Downs, and Yang Yang.
\newblock Firebolt: Weak supervision under weaker assumptions.
\newblock In \emph{International Conference on Artificial Intelligence and Statistics}, pp.\  8214--8259. PMLR, 2022.

\bibitem[Mazzetto et~al.(2021)Mazzetto, Cousins, Sam, Bach, and Upfal]{mazzetto2021adversarial}
Alessio Mazzetto, Cyrus Cousins, Dylan Sam, Stephen~H Bach, and Eli Upfal.
\newblock Adversarial multi class learning under weak supervision with performance guarantees.
\newblock In \emph{International Conference on Machine Learning}, pp.\  7534--7543. PMLR, 2021.

\bibitem[Medlock \& Briscoe(2007)Medlock and Briscoe]{medlock2007weakly}
Ben Medlock and Ted Briscoe.
\newblock Weakly supervised learning for hedge classification in scientific literature.
\newblock In \emph{Proceedings of the 45th annual meeting of the association of computational linguistics}, pp.\  992--999, 2007.

\bibitem[Mintz et~al.(2009)Mintz, Bills, Snow, and Jurafsky]{mintz2009distant}
Mike Mintz, Steven Bills, Rion Snow, and Dan Jurafsky.
\newblock Distant supervision for relation extraction without labeled data.
\newblock In \emph{Proceedings of the Joint Conference of the 47th Annual Meeting of the ACL and the 4th International Joint Conference on Natural Language Processing of the AFNLP}, pp.\  1003--1011, 2009.

\bibitem[Nikolenko(2021)]{nikolenko2021synthetic}
Sergey~I Nikolenko.
\newblock \emph{Synthetic data for deep learning}, volume 174.
\newblock Springer, 2021.

\bibitem[Ratner et~al.(2018)Ratner, Hancock, Dunnmon, Goldman, and R{\'e}]{ratner2018snorkel}
Alex Ratner, Braden Hancock, Jared Dunnmon, Roger Goldman, and Christopher R{\'e}.
\newblock Snorkel metal: Weak supervision for multi-task learning.
\newblock In \emph{Proceedings of the Second Workshop on Data Management for End-To-End Machine Learning}, pp.\  1--4, 2018.

\bibitem[Ratner et~al.(2019)Ratner, Hancock, Dunnmon, Sala, Pandey, and R{\'e}]{ratner2019training}
Alexander Ratner, Braden Hancock, Jared Dunnmon, Frederic Sala, Shreyash Pandey, and Christopher R{\'e}.
\newblock Training complex models with multi-task weak supervision.
\newblock In \emph{Proceedings of the AAAI Conference on Artificial Intelligence}, volume~33, pp.\  4763--4771, 2019.

\bibitem[Ratner et~al.(2016)Ratner, De~Sa, Wu, Selsam, and R{\'e}]{ratner2016data}
Alexander~J Ratner, Christopher~M De~Sa, Sen Wu, Daniel Selsam, and Christopher R{\'e}.
\newblock Data programming: Creating large training sets, quickly.
\newblock \emph{Advances in neural information processing systems}, 29, 2016.

\bibitem[Settles(1994)]{settles1994active}
Burr Settles.
\newblock Active learning literature survey.
\newblock \emph{Machine Learning}, 15\penalty0 (2):\penalty0 201--221, 1994.

\bibitem[Shin et~al.(2015)Shin, Wu, Wang, De~Sa, Zhang, and R{\'e}]{shin2015incremental}
Jaeho Shin, Sen Wu, Feiran Wang, Christopher De~Sa, Ce~Zhang, and Christopher R{\'e}.
\newblock Incremental knowledge base construction using deepdive.
\newblock In \emph{Proceedings of the VLDB Endowment International Conference on Very Large Data Bases}, volume~8, pp.\  1310. NIH Public Access, 2015.

\bibitem[Varadi et~al.(2022)Varadi, Anyango, Deshpande, Nair, Natassia, Yordanova, Yuan, Stroe, Wood, Laydon, et~al.]{varadi2022alphafold}
Mihaly Varadi, Stephen Anyango, Mandar Deshpande, Sreenath Nair, Cindy Natassia, Galabina Yordanova, David Yuan, Oana Stroe, Gemma Wood, Agata Laydon, et~al.
\newblock Alphafold protein structure database: massively expanding the structural coverage of protein-sequence space with high-accuracy models.
\newblock \emph{Nucleic acids research}, 50\penalty0 (D1):\penalty0 D439--D444, 2022.

\bibitem[Weiss et~al.(2016)Weiss, Khoshgoftaar, and Wang]{weiss2016survey}
Karl Weiss, Taghi~M Khoshgoftaar, and DingDing Wang.
\newblock A survey of transfer learning.
\newblock \emph{Journal of Big data}, 3\penalty0 (1):\penalty0 1--40, 2016.

\bibitem[Wolf et~al.(2020)Wolf, Debut, Sanh, Chaumond, Delangue, Moi, Cistac, Rault, Louf, Funtowicz, et~al.]{wolf2020transformers}
Thomas Wolf, Lysandre Debut, Victor Sanh, Julien Chaumond, Clement Delangue, Anthony Moi, Pierric Cistac, Tim Rault, R{\'e}mi Louf, Morgan Funtowicz, et~al.
\newblock Transformers: State-of-the-art natural language processing.
\newblock In \emph{Proceedings of the 2020 conference on empirical methods in natural language processing: system demonstrations}, pp.\  38--45, 2020.

\bibitem[Wu et~al.(2022)Wu, Chen, Zhang, and Chu]{wu2022learning}
Renzhi Wu, Shen-En Chen, Jieyu Zhang, and Xu~Chu.
\newblock Learning hyper label model for programmatic weak supervision.
\newblock In \emph{The Eleventh International Conference on Learning Representations}, 2022.

\bibitem[Wu et~al.(2023)Wu, Bendeck, Chu, and He]{wu2023ground}
Renzhi Wu, Alexander Bendeck, Xu~Chu, and Yeye He.
\newblock Ground truth inference for weakly supervised entity matching.
\newblock \emph{Proceedings of the ACM on Management of Data}, 1\penalty0 (1):\penalty0 1--28, 2023.

\bibitem[Xiaojin(2006)]{xiaojin2006semi}
Zhu Xiaojin.
\newblock Semi-supervised learning literature sur-vey.
\newblock \emph{Semi-Supervised Learning Literature Sur-vey, Technical report, Computer Sciences, University of Wisconsin-Madisoa}, 2006.

\bibitem[Yu et~al.(2022)Yu, Ding, and Bach]{yu2022learning}
Peilin Yu, Tiffany Ding, and Stephen~H Bach.
\newblock Learning from multiple noisy partial labelers.
\newblock In \emph{International Conference on Artificial Intelligence and Statistics}, pp.\  11072--11095. PMLR, 2022.

\bibitem[Zhang et~al.(2021)Zhang, Yu, Li, Wang, Yang, Yang, and Ratner]{zhang2021wrench}
Jieyu Zhang, Yue Yu, Yinghao Li, Yujing Wang, Yaming Yang, Mao Yang, and Alexander Ratner.
\newblock Wrench: A comprehensive benchmark for weak supervision.
\newblock \emph{arXiv preprint arXiv:2109.11377}, 2021.

\bibitem[Zhang et~al.(2022)Zhang, Hsieh, Yu, Zhang, and Ratner]{zhang2022survey}
Jieyu Zhang, Cheng-Yu Hsieh, Yue Yu, Chao Zhang, and Alexander Ratner.
\newblock A survey on programmatic weak supervision.
\newblock \emph{arXiv preprint arXiv:2202.05433}, 2022.

\end{thebibliography}
\bibliographystyle{iclr2023_conference}

\appendix
\end{document}